\documentclass[10pt,twocolumn,letterpaper]{article}

\usepackage[pagenumbers]{wacv}

\definecolor{wacvblue}{rgb}{0.21,0.49,0.74}
\usepackage[pagebackref,breaklinks,colorlinks,allcolors=wacvblue]{hyperref}
 \usepackage{comment}
\def\wacvPaperID{2962} 
\def\confName{WACV}
\def\confYear{2027}

\title{Foresight at the Event Boundary: Evaluating\\ Physical Prediction in Video World Models}

\author{
Estela Monserrat Arriaga Santana$^{1}$ \quad 
Julian Rosas Scull$^{1}$ \quad
Eh\'ecatl Sacamch'en N\'u\~nez Rico$^{1}$ \quad \\
Hugo Jair Escalante$^{2}$\\[2pt]
$^{1}$National Autonomous University of Mexico, Mexico\\
$^{2}$University of Texas at El Paso, USA\\[2pt]
{\tt\small
mon.arriaga.santana@gmail.com \quad
julian.rosas@ciencias.unam.mx}\\
{\tt\small
ehecatl564@gmail.com \quad
hescalantebal@utep.edu}
}

\begin{document}
\maketitle

\begin{abstract}
Video world models are largely regarded as predictive models of the physical world, and are therefore expected to anticipate the consequences of observed events. However, up to now, evaluation has mainly focused on reference similarity, physical-law consistency, or judgment plausibility, estimating anticipation only indirectly. We address this straightforwardly: when a release or impact has just occurred but its consequence is withheld, can a world model anticipate what should happen next? We introduce an event-anchored evaluation based on $62$ controlled real-world free-fall recordings and $124$ clips spanning three object types, with fine-grained release and impact annotations and ground-truth trajectories. The protocol separates consequence production, temporal placement, and physical realization. Across six contemporary video generation and world models, our findings reveal interesting failure patterns: Runway and Veo produce release and subsequent impact events at rates above $93\%$ but often initiate them substantially late, whereas Cosmos-Predict-2.5 and MAGI-1 frequently preserve the pre-event state and produce little or no measurable consequence. Among measurable falls, plausible timing does not necessarily imply physically consistent motion. We further conduct a $15$-participant, $20$-condition human study in which participants describe the expected consequence from a single event-anchored frame and draw its trajectory, allowing us to contrast human and model performance. Human predictions favor the recorded future in aggregate while revealing genuine ambiguity among plausible continuations. Trajectory analysis further shows that whether a model produces measurable motion must be separated from how accurately that motion is realized. Overall, physical foresight emerges as a sequence of distinct challenges: initiating a consequence, anchoring it in time, and realizing its motion.
\end{abstract}
\input
\section{Introduction}
\label{sec:intro}

Video world models are increasingly framed as predictive models of the physical
world. Under this view, a model should not only generate visually plausible
motion, but also continue the consequence of an event that has already been
observed: an unsupported object should fall after release, and an object that
has just collided with a surface should react accordingly.

Existing evaluations largely ask whether a generated continuation resembles a
recorded future or satisfies physical constraints
\cite{motamed2026physicsiq,zhang2025morpheus,upadhyay2026worldbench}. These
criteria can miss an earlier failure: the model may never initiate the expected
consequence at all. We therefore separate three questions: whether the observed
event is carried through, whether the resulting motion is physically plausible,
and whether the expected consequence is inferable from the available evidence.

We study these questions using controlled real-world free-fall recordings of
three objects, annotated at release and impact and accompanied by measured
trajectories. Prediction ends exactly at one of these physical events, and
everything that follows is withheld. We evaluate six models---Cosmos-3~\cite{nvidia2026cosmos3},
Cosmos-Predict-2.5~\cite{ali2025cosmospredict25},
MAGI-1~\cite{teng2025magi1},
PhyWorld~\cite{zhao2026phyworld},
Runway Gen-4.5~\cite{runway2025gen45}, and
Google Veo 3.1~\cite{google2025veo31}---and complement the model evaluation
with a human study. Because the recorded continuation represents only one possible future compatible with the observed event, human predictions provide a reference for
which consequences and trajectories are reasonably anticipated from the available evidence.

The results reveal qualitatively different failure modes. Runway and Veo usually
produce the expected release consequence, but often place it substantially too
late; Cosmos-Predict-2.5 and MAGI-1 frequently fail to initiate the consequence
at all, while Cosmos-3 and PhyWorld lie between these regimes. Once measurable
free-fall motion is produced, its timing can be close to the real fall even when
the event onset is misplaced. Human predictions further show that the recorded
future receives the strongest aggregate support from minimal evidence, while
also revealing genuine ambiguity among plausible continuations.

The contributions of this work are as follows: 
\begin{itemize}
    \item an event-anchored, outcome-blind protocol that separates consequence
    production, temporal placement, and physical realization;
    \item a new annotated real-world free-fall resource comprising $62$
    recordings and $124$ event-anchored clips, with release, impact, and rest
    events together with metric object trajectories;
    \item an evaluation of six recent video generation and world models; and
    \item a matched human study of open-ended consequence and trajectory
    prediction from the same event anchors.
\end{itemize}

\section{Related Work}
\label{sec:related_work}

Evaluations of generated video typically rely on human/VLM judgments,
comparison with a reference future, or explicit tests of physical laws.
PhyGround and Physion-Eval provide semantic judgments of physical failures
\cite{lin2026phyground,zhang2026physioneval}, while Physics-IQ compares generated
motion with real experimental recordings
\cite{motamed2026physicsiq,radsch2026physicsiqverified}. Reference-based
evaluation measures agreement with an observed future, but may penalize
alternative physically valid continuations when the initial state is not fully
determined.

Law-based benchmarks instead assess physical admissibility. Morpheus evaluates
trajectory-level equations and invariants such as acceleration, energy,
momentum, and period \cite{zhang2025morpheus}, while WorldBench combines
concept-specific evaluation with estimation of physical parameters such as
gravity and friction \cite{upadhyay2026worldbench}. These approaches measure
whether generated motion follows expected physics, but they do not specifically
isolate the response to an observed physical event. In particular, they do not
directly measure whether a consequence begins after a release or impact, how
long that onset is delayed, or whether the response is suppressed altogether.
Physion and Physion++ also define explicit prediction boundaries
\cite{bear2021physion,tung2023physionpp}, but these boundaries are not used as
event anchors for measuring the subsequent physical response.

Our protocol focuses on this complementary event-level setting. Observation
ends at an annotated release or impact in controlled real-world free-fall
experiments, and the subsequent consequence is withheld. We evaluate whether
the expected response occurs, when it begins, and whether the resulting motion
is physically plausible. We further compare these continuations with open-ended
human predictions from the same event anchors. Unlike the predefined contact
judgments used in Physion and Physion++, participants freely describe and draw
the expected continuation, following a broader tradition of using trajectory
drawings to probe intuitive physics
\cite{bear2021physion,tung2023physionpp,mccloskey1980curvilinear}.
\section{Dataset}
\label{sec:dataset}

\begin{figure}[h]
    \centering
    \includegraphics[width=0.9\columnwidth]{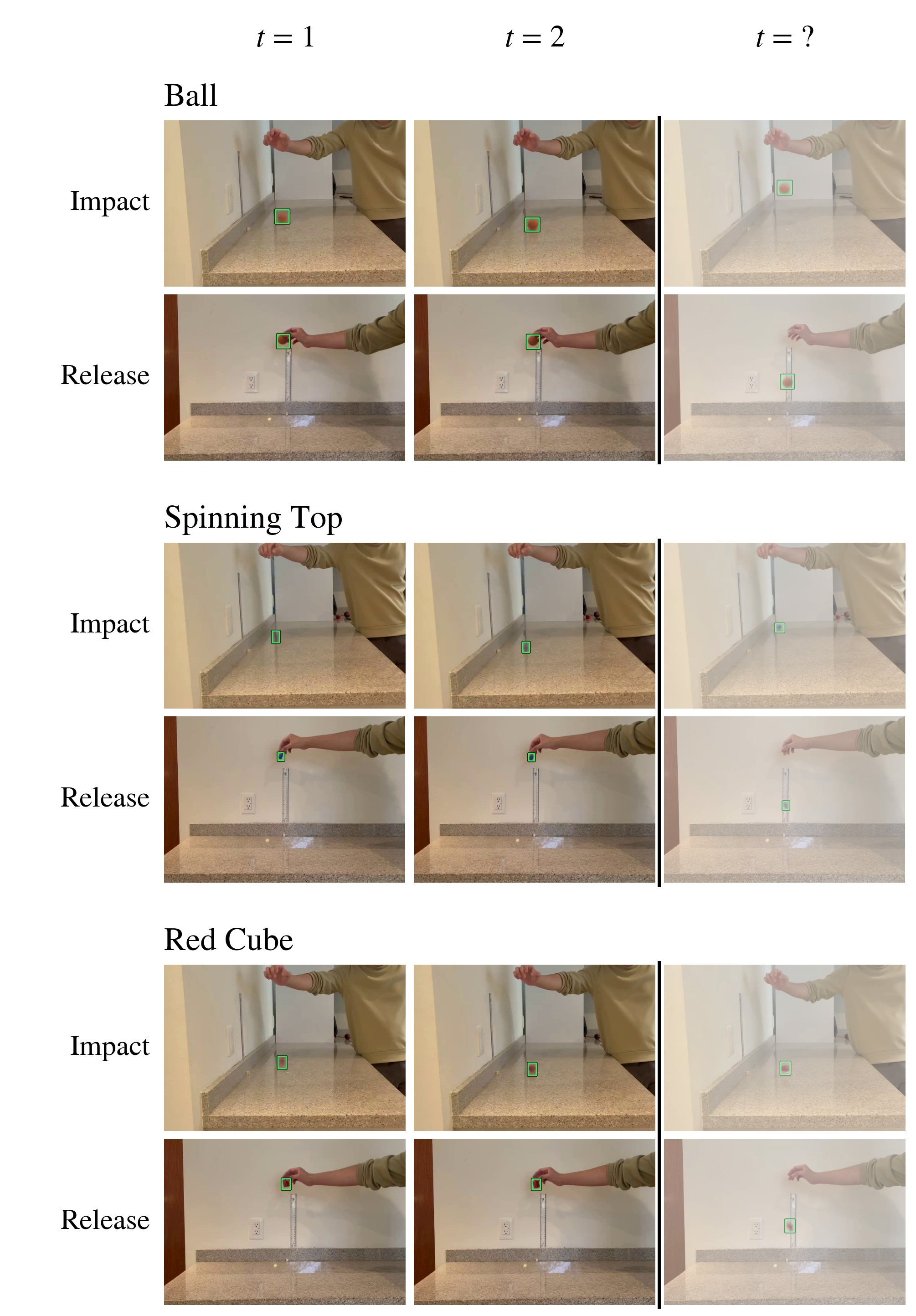}
    \caption{
    Event-anchored evaluation clips. The context ends at the annotated
    release or impact; subsequent frames are withheld from the model.
    Examples show the three evaluated objects.
    }
    \label{fig:dataset-event-anchors}
\end{figure}

\paragraph{Source recordings and event clips.}
We use $62$ real free-fall recordings captured at $832\times464$ and
$59.9$\,fps under fixed indoor conditions, covering three objects---a ball,
a red cube, and a spinning top---and two camera viewpoints. The objects were
chosen to provide contrasting geometries and contact profiles within the same
controlled free-fall setting: although all three are subject to the same
gravitational acceleration during the fall, their shape and mass distribution
support substantially different post-contact responses, ranging from
bouncing or rolling to face-supported and rotational motion. This allows us
to test a common event-conditioned prediction problem without restricting the
evaluation to a single object geometry. Each recording is split into a release-anchored and an impact-anchored clip, yielding $124$ event clips (Fig.~\ref{fig:dataset-event-anchors}). Each clip provides $33$ context frames ($0.55$\,s), ending at the annotated anchor event, and the prompt describes the observed setup without revealing the future
trajectory or outcome.

\paragraph{Annotations.}
The real recordings were annotated in CVAT~\cite{cvat} with object locations
and frame-level release, impact, and rest events. We additionally tracked the
objects in Tracker~\cite{brown2026tracker}, using the per-video spatial
calibration to obtain metric trajectories for the physics-based measurements.
Generated videos were annotated only in CVAT, using the same object tracks and
event markers, together with failure-mode labels such as suppressed motion,
disappearance, duplication, and outcome substitution. Four annotators labeled
the real and generated videos, with each video assigned to one annotator.

\paragraph{Generated and evaluated clips.}
We evaluate Cosmos-3, Cosmos-Predict-2.5, MAGI-1, PhyWorld, Runway, and Veo,
producing $743$ videos in total ($124$ per model except Veo, which failed on one
clip). For manual evaluation, we use the $123$ event clips available for all six
models and sample a balanced subset over object, viewpoint, and anchor event.
The $3\times2\times2=12$ strata contribute eight clips each, giving $96$ event
clips per model and $576$ annotated generated videos in total.

\section{Proposed Evaluation Protocol}
Our protocol evaluates physical foresight along three complementary dimensions:
whether an observed event is carried through to its expected consequence,
whether the resulting motion is physically plausible, and whether the generated
future aligns with human expectations under minimal visual evidence. The
physical-realization criteria follow established physics-based evaluation
practice, while our event-level consequence measurement and open-ended human
reference target complementary aspects not isolated by these evaluations.

\subsection{Measuring consequence production}
\label{sec:method-consequence}

Under the considered scenario, the prompt and the context video specify which event has just occurred and withhold everything that follows from it.  The evaluation protocol  establishes whether the generated continuation contains the events entailed by that anchor event. It does not assess how accurately those events are rendered: no criterion of physical plausibility, trajectory accuracy or timing is imposed, so that a model is credited with the consequence whenever it produces it at all.

\paragraph{Release condition.}
The context ends at the instant the object is let go. The entailed continuation
comprises a \emph{release}, a \emph{subsequent impact}, and the \emph{object coming to rest} after
moving. The presence of each of the three markers is recorded independently.
Release and impact constitute the substantive tests; the rest marker carries
weaker evidential weight, since a clip may terminate before the object settles.

\paragraph{Impact condition.}
The context ends at the instant of contact. The requirement is that the model,
from the prompt and the reference frames, infer what follows: the object must
react rather than remain frozen. The same three markers are annotated as in the
release condition, and they are used to characterise what the model produced---in
particular whether it substituted a fresh release for the expected continuation,
and whether the object was brought to rest.

The markers do not, however, establish whether the object moved. The impact
marker coincides with the anchor in this condition, since the contact is supplied
to the model rather than predicted by it, and the rest marker records that the
object is at rest irrespective of whether it had previously been in motion.
Motion is consequently determined from the annotated trajectory: an object counts
as having moved when the centre of its bounding box departs from its initial
position by more than half the object's own size. Displacement is expressed in
object diameters---the mean diagonal of the bounding box---rather than in pixels,
since each model generates at its native resolution ($832\times480$ to
$1280\times720$) and a pixel threshold would not be comparable across models.

\paragraph{Scope of the measurement.}

Event rates are computed over the full generated portion of each clip, without
restricting when the event must occur. For models that replay conditioning
frames at the start of their output (Cosmos-3: $13$; Cosmos-Predict-2.5: $5$;
PhyWorld: $9$), evaluation begins at the first newly generated frame, since the
replayed frames are part of the input and already contain the anchor event.
MAGI-1, Runway, and Veo require no such exclusion.
Release markers occurring within replayed conditioning frames are ignored;
only releases occurring after the first genuinely generated frame are treated
as generated re-release events.

\begin{figure*}[h]
    \centering
    \includegraphics[width=0.65\textwidth]{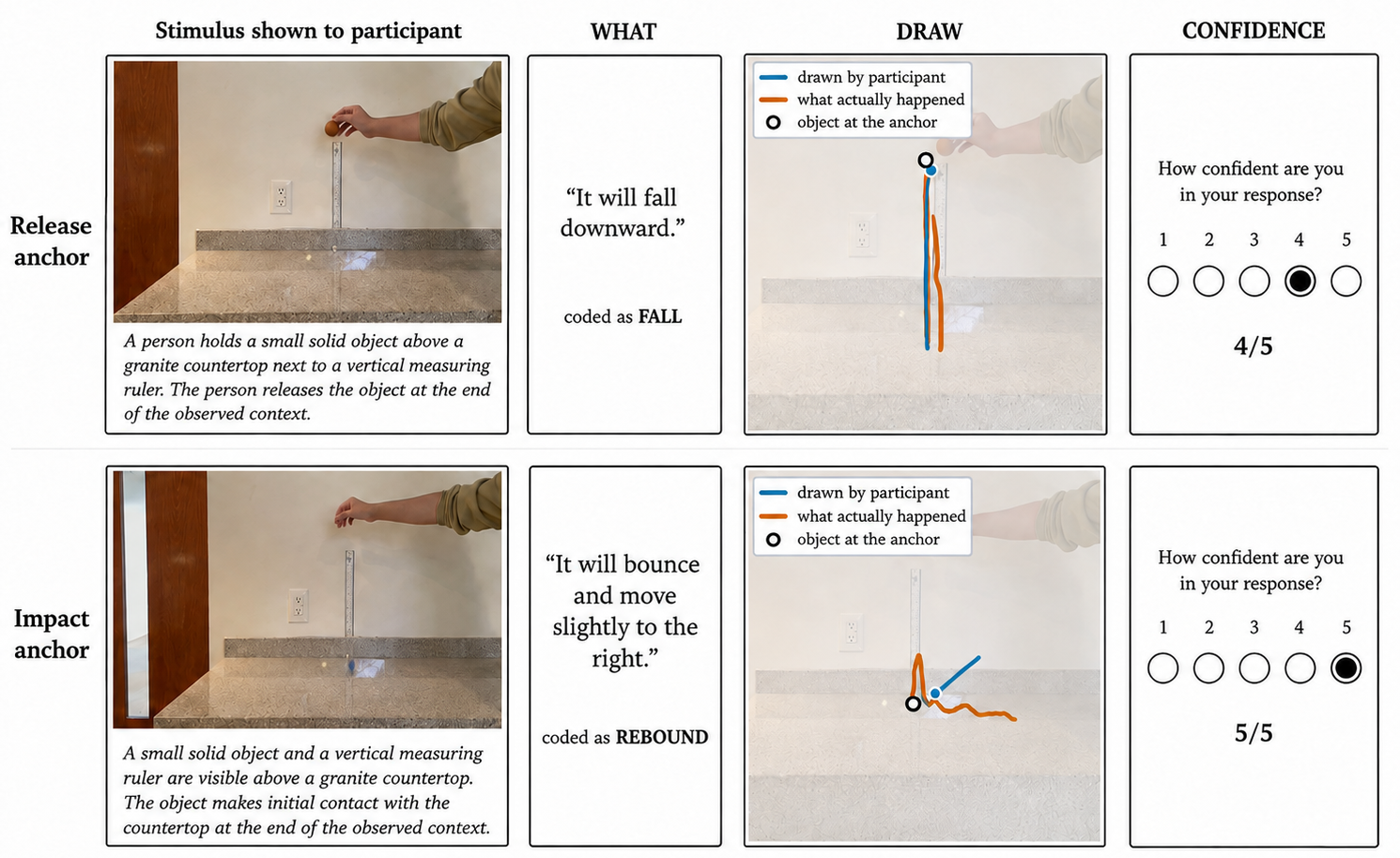}
    \caption{
    Human prediction protocol. Participants observe an anchor frame and an
    outcome-blind event description, then provide a textual prediction
    (\emph{WHAT}), a trajectory drawing (\emph{DRAW}), and a confidence score.
    The recorded trajectory is used only for evaluation and is never shown to
    participants.
    }
    \label{fig:human-study-protocol}
\end{figure*}

\subsection{Measuring physical realization}

Following established physics-based evaluation
practice~\cite{motamed2026physicsiq,zhang2025morpheus,
upadhyay2026worldbench}, we assess measurable free-fall motion along three
criteria. We compare the annotated release-to-impact duration with the
theoretical time $t=\sqrt{2h/g}$, using $g=9.81\,\mathrm{m/s^2}$ and a
one-frame tolerance determined by the video's frame rate. We additionally
measure the horizontal displacement of the object's bounding-box center from
its release position, allowing a $10$ px tolerance, and estimate
$g_{\mathrm{model}}$ by fitting the tracked vertical trajectory to a
constant-acceleration model. These measurements are computed only for
continuations with identifiable release and impact events. Full derivations,
fitting details, and implementation choices are provided in the supplementary
material.

\subsection{Human foresight reference}
\label{sec:method-human}

\paragraph{Human prediction study.}
We evaluate whether the expected continuation can be inferred from minimal
visual evidence through a separate human study. The human prediction protocol is summarized in
Fig.~\ref{fig:human-study-protocol}. Fifteen participants complete $20$
free-fall conditions ($11$ release and $9$ impact), yielding $300$ responses. For each condition, participants see only the final anchor frame
together with the same outcome-blind event description used for generation.
They predict what happens next in free text (\emph{WHAT}), draw the expected
object trajectory (\emph{DRAW}), and report confidence on a five-point scale.
The conditions span three objects, two viewpoints, and different release
heights.

\paragraph{WHAT: predicted consequence.}
Free-text responses are assigned a single code corresponding to the first
physical consequence explicitly predicted by the participant
\cite{hsieh2005three,mayring2000qualitative}. Categories include
\textsc{fall}, \textsc{rebound}, \textsc{surface-motion}, \textsc{no-motion},
\textsc{contact}, \textsc{rotate}, \textsc{break}, and \textsc{other}; we do
not infer unstated intermediate events, so ``fall and bounce'' is coded as
\textsc{fall}. To compare humans with models and the recorded future, generated
and real trajectories are mapped to the same consequence space. Starting at the
anchor, the first displacement exceeding $0.5d_0$, where $d_0$ is the object's
bounding-box diagonal, is classified as \textsc{fall}, \textsc{rebound}, or
\textsc{surface-motion} according to whether motion is predominantly downward,
upward, or horizontal; trajectories that never reach the threshold are coded as
\textsc{no-motion}. We then report the fraction of participants predicting the
same consequence as each model and the recorded future.

\paragraph{DRAW: predicted trajectory.}
Human drawings, model trajectories, and recorded trajectories are translated to
the anchor position and normalized by $d_0$, so distances are expressed in
object-size units rather than pixels. Curves are resampled to $30$ points and
compared over the corresponding event interval: release to first contact for
release trials, and the initial post-impact response for impact trials.
Trajectories moving less than $0.5d_0$ are treated as degenerate and receive no
geometric score; for the remaining trajectories, we use discrete Fr\'echet
distance~\cite{eiter1994computing}, which compares path geometry without
requiring synchronized timestamps. We report human--human ($H$--$H$),
human--ground-truth ($H$--$GT$), model--human ($M$--$H$), and
model--ground-truth ($M$--$GT$) comparisons.

\paragraph{Conditioning asymmetry.}
To define a human reference under the minimum visual evidence available across
the evaluated models, participants receive only the anchor frame together with
the outcome-blind event description. This matches the conditioning of the
single-frame models, Runway and Veo. The video-conditioned models receive
additional temporal context according to their native input requirements.
Accordingly, comparisons with Runway and Veo are matched in visual evidence,
whereas comparisons with the video-conditioned models should be interpreted as
alignment with a minimal-evidence human reference rather than as a controlled
test of context length.

\section{Experimental Settings}
\label{sec:experimental-settings}
We evaluate six recent video generation and world models:
Cosmos3-Nano~\cite{nvidia2026cosmos3},
Cosmos-Predict2.5-2B~\cite{ali2025cosmospredict25},
MAGI-1-4.5B-distill~\cite{teng2025magi1},
PhyWorld~\cite{zhao2026phyworld},
Runway Gen-4.5~\cite{runway2025gen45}, and
Google Veo 3.1~\cite{google2025veo31}.
Each open-weight model was self-hosted on a single NVIDIA A100 80\,GB GPU
through RunPod,\footnote{\url{https://console.runpod.io/}} while Runway and
Veo were accessed through Replicate using the
\texttt{runwayml/gen-4.5}\footnote{\url{https://replicate.com/runwayml/gen-4.5}}
and \texttt{google/veo-3.1}\footnote{\url{https://replicate.com/google/veo-3.1}}
endpoints. We preserve each model's native conditioning modality and recommended
operating point whenever compatible with our protocol; exact inference
configurations and deviations from released decoding settings are reported in
the supplementary material.

\paragraph{Models and conditioning.}
The models receive the temporal evidence supported by their native
conditioning interfaces (Table~\ref{tab:model-conditioning}). Cosmos3-Nano,
MAGI-1, and PhyWorld receive the complete $0.55$\,s event-ending context,
resampled to their required frame rate. Cosmos-Predict2.5-2B uses its default
five-frame Video2World conditioning ($\approx0.31$\,s), while Runway Gen-4.5
and Veo 3.1 operate from the anchor frame alone. In all cases, conditioning
terminates at the annotated release or impact and contains no post-anchor
information.

\begin{table}[t]
\centering
\small
\setlength{\tabcolsep}{5pt}
\begin{tabular}{lccc}
\toprule
Model & Access & Mode & Visual context \\
\midrule
Cosmos3-Nano         & Open & V2V & $0.55$\,s \\
Cosmos-Predict2.5-2B & Open & V2V & 5 frames ($\approx0.31$\,s) \\
MAGI-1-4.5B-distill  & Open & V2V & $0.55$\,s \\
PhyWorld             & Open & V2V & $0.55$\,s \\
Runway Gen-4.5       & Closed & I2V & 1 frame \\
Google Veo 3.1       & Closed & I2V & 1 frame \\
\bottomrule
\end{tabular}
\caption{Access and visual conditioning of the evaluated models. V2V and I2V
denote video- and image-conditioned generation, respectively.}
\label{tab:model-conditioning}
\end{table}

\paragraph{Prompting.}
Following Physics-IQ Verified~\cite{radsch2026physicsiqverified}, prompts
describe the setup, camera, and anchor event while withholding the subsequent
motion, trajectory, and outcome. Wording is adapted to each model's interface
while preserving the same event-level information. Where supported, auxiliary
constraints discourage camera motion and generation artifacts; for Runway,
which lacks a negative-prompt field, these are expressed positively.

\section{Results}
We first examine whether and when models produce the expected event
consequence, and then compare their predicted consequences and trajectories
with human expectations under minimal visual evidence.

\subsection{Does the model produce the consequence?}
\label{sec:results-consequence}

\begin{table*}[h]
\centering
\small
\setlength{\tabcolsep}{7pt}
\begin{tabular}{lccccc}
\toprule
& \multicolumn{3}{c}{Release condition} & & Impact condition \\
\cmidrule(lr){2-4}\cmidrule(l){6-6}
Model & Release & Impact & Rest & & Object moved \\
\midrule
Cosmos3-Nano
& 33.3 [22--47]
& 33.3 [22--47]
& 62.5 [48--75]
&& \textbf{100.0} [93--100] \\

Cosmos-P2.5
& \phantom{0}0.0 [0--7]
& \phantom{0}0.0 [0--7]
& 47.9 [34--62]
&& 27.1 [17--41] \\

MAGI-1
& \phantom{0}2.1 [0--11]
& \phantom{0}0.0 [0--7]
& 56.2 [42--69]
&& 36.2 [24--50] \\

PhyWorld
& 33.3 [22--47]
& 31.2 [20--45]
& 64.6 [50--77]
&& 72.9 [59--83] \\

Runway
& \textbf{95.8} [86--99]
& \textbf{93.8} [83--98]
& \textbf{95.8} [86--99]
&& 85.4 [73--93] \\

Veo
& \textbf{95.8} [86--99]
& \textbf{93.8} [83--98]
& \textbf{95.8} [86--99]
&& 81.2 [68--90] \\
\bottomrule
\end{tabular}

\caption{
Occurrence rates of event-level consequences. Under the release condition,
columns indicate whether release, subsequent impact, and rest occur; under the
impact condition, whether the object moves after the observed impact. Values are
percentages with 95\% Wilson confidence intervals. There are $n=48$ clips per
model and condition, except MAGI-1 impact ($n=47$) due to one missing object
annotation.
}
\label{tab:consequence}
\end{table*}

Table~\ref{tab:consequence} reveals a sharp difference in consequence
production. Runway and Veo carry the release through to a subsequent impact in
over $93\%$ of clips. Cosmos-3 and PhyWorld do so in roughly one third, whereas
Cosmos-Predict-2.5 never generates a release and MAGI-1 does so only once.
For Cosmos-3 and PhyWorld, impact rates closely track release rates
($33.3$ vs.\ $33.3$ and $33.3$ vs.\ $31.2$), indicating that most failures
occur at the first step: once the object is released, the remaining consequence
usually follows.

The rest marker should be interpreted separately. Cosmos-Predict-2.5 and MAGI-1
are labeled at rest in $47.9\%$ and $56.2\%$ of clips despite almost never
releasing the object, because \emph{rest} indicates only that the object is
stationary, not that it came to rest after moving. We therefore do not interpret
this marker alone as evidence of a completed consequence.

The impact condition exposes a related failure. Cosmos-3 moves the object in
every clip, while Runway, Veo, and PhyWorld do so in most cases.
Cosmos-Predict-2.5 and MAGI-1 instead leave the object frozen in $72.9\%$ and
$63.8\%$ of clips, respectively, even though the collision itself is already
visible in the conditioning input.

A second failure mode is \emph{event replay}: rather than continuing from the observed impact, some models regenerate an earlier stage of the event. After excluding releases contained in replayed conditioning frames, Veo generates a new release in $17$ impact-conditioned clips and Runway in $4$. Several Veo clips contain multiple re-releases, indicating repetition of the causal event  rather than continuation from it.

Together, the two conditions distinguish two forms of consequence failure:
suppression of the expected response, most evident for Cosmos-Predict-2.5 and
MAGI-1, and replay of an earlier causal stage, most evident for Veo.
\paragraph{When the consequence is placed.}
High event-occurrence rates can hide substantial timing errors. As shown in
Table~\ref{tab:timing}, Runway and Veo release the object after median onset
times of $1.73$\,s and $0.94$\,s, compared with $0.23$\,s for Cosmos-3 and
$0.22$\,s for PhyWorld. Restricting evaluation to the first generated second
therefore reduces the release rate from $95.8\%$ to $22.9\%$ for Runway and to
$54.2\%$ for Veo, while leaving the other models largely unchanged.

The delay occurs mainly before the release. Once falling begins, Runway
($0.25$\,s) and Veo ($0.21$\,s) are close to the real release-to-impact duration
of $0.20$\,s; Cosmos-3 and PhyWorld yield $0.29$\,s and $0.25$\,s,
respectively. Thus, Runway and Veo often produce the expected consequence but
place it too late, whereas Cosmos-Predict-2.5 and MAGI-1 rarely initiate it at
all. Because Runway and Veo receive only the final context frame, their delayed
onset may partly reflect the absence of temporal context rather than model
quality alone.

\begin{table}[h]
\centering
\small
\begin{tabular}{lcccc}
\toprule
& Releases & Onset & Fall & Settling \\
\midrule
Real recordings    & 61 & ---  & 0.20 & 1.12 \\
\midrule
Cosmos-3           & 16 & 0.23 & 0.29 & 0.25 \\
Cosmos-Predict-2.5 & \phantom{0}0 & ---  & ---  & ---  \\
MAGI-1             & \phantom{0}1 & ---  & ---  & ---  \\
PhyWorld           & 16 & 0.22 & 0.25 & 0.31 \\
Runway             & 46 & \textbf{1.73} & 0.25 & 2.29 \\
Veo                & 46 & \textbf{0.94} & 0.21 & 2.25 \\
\bottomrule
\end{tabular}
\caption{Median event timing in seconds. Onset is measured from the first
generated frame to release, fall from release to impact, and settling from
impact to rest. Releases gives the number of clips contributing to the onset
estimate; later intervals may contain fewer valid clips.}
\label{tab:timing}
\end{table}

\subsection{Is the resulting motion physically plausible?}

Table~\ref{tab:physical-realization} summarizes performance across the three
classical-mechanics-based criteria described above. The first two columns report
the fraction of videos satisfying the expected temporal and spatial properties
of free fall, respectively, while the last reports the mean estimated
gravitational acceleration and its standard deviation. These measurements are
computed only on release-conditioned continuations in which a generated release
and subsequent impact can be identified. Re-release events generated after an
impact anchor are treated as event-replay failures
(Sec.~\ref{sec:results-consequence}) and excluded from the primary physics
analysis.

The results reveal substantial differences across models and show that
satisfying one physical criterion does not necessarily imply consistency with
the others. PhyWorld, for example, produces plausible fall durations in $50\%$
of the analyzed videos and exhibits no measurable horizontal deviation in
$36\%$ of them. However, its estimated gravitational acceleration is
$\bar{g}_{\mathrm{model}}=3.36\pm3.12\,\mathrm{m/s^2}$, considerably below the
expected value of $9.81\,\mathrm{m/s^2}$. Thus, a generated trajectory may
satisfy some quantitative physical criteria while remaining inconsistent with
the expected dynamics.

Cosmos-3 provides the closest mean estimate to the expected gravitational
acceleration, with
$\bar{g}_{\mathrm{model}}=8.09\pm11.94\,\mathrm{m/s^2}$. Nevertheless, only
$46\%$ of its videos satisfy the temporal criterion and $8\%$ satisfy the
criterion for the absence of horizontal deviation. The large standard
deviation further indicates substantial variability across samples, so the
agreement of the mean acceleration with the physical reference is not
representative of all generated continuations.

Runway and Veo yield lower mean accelerations of
$5.35\pm5.67\,\mathrm{m/s^2}$ and $4.47\pm4.85\,\mathrm{m/s^2}$,
respectively, and show no measurable horizontal deviation in only $4\%$ and
$2\%$ of videos. MAGI-1 and Cosmos-Predict-2.5 produce no release-conditioned
continuations for which both release and impact can be reliably identified, so
no physical measurements are reported for these models. Overall, the
disagreement between fall duration, horizontal consistency, and estimated
acceleration shows that physical plausibility cannot be adequately
characterized by a single measurement. Here, fall duration captures temporal
consistency, horizontal deviation captures spatial consistency, and the
estimated acceleration provides a direct measure of the generated dynamics.

\begin{table*}[h]
\centering
\small
\begin{tabular}{lccc}
\toprule
Model & Plausible Fall Time & No Horizontal Deviation & $\bar{g}_{\mathrm{model}}\pm\sigma\;[\mathrm{m/s^2}]$ \\

\midrule
Cosmos-3           & 0.46 & 0.08 & 8.09$\pm$11.94  \\
Cosmos-Predict-2.5 & --- & ---  & ---\\
MAGI-1             & --- & ---  & ---   \\
PhyWorld           & 0.50 & 0.36 & 3.36$\pm$3.12  \\
Runway             & 0.30 & 0.04 & 5.35$\pm$ 5.67\\
Veo                & 0.27 & 0.02 & 4.47$\pm$ 4.85  \\
\bottomrule
\end{tabular}
\caption{Fraction of videos that satisfied the classical-mechanics-based benchmarks, together with the estimated accelerations. Only videos with a release anchor exhibiting both a successful release and an impact were considered.}
\label{tab:physical-realization}
\end{table*}

\subsection{Do model predictions align with human foresight?}
\label{sec:results-human}

We compare models and the recorded future with human predictions in two spaces:
\emph{WHAT}, measuring human support for the first predicted consequence, and
\emph{DRAW}, measuring trajectory production and, conditional on motion,
geometry relative to humans and ground truth.

\paragraph{WHAT: Which consequence is anticipated?}
At the human level, open-ended responses remained distributed, with
primary-event entropy of $1.42$ bits for release and $1.91$ bits for impact,
indicating greater disagreement after impact. For human--model alignment,
Table~\ref{tab:human-support} reports, for each model and the recorded future,
the mean fraction of participants predicting the same first consequence. The
recorded continuation receives the greatest aggregate support in both
conditions: $0.61$ for release and $0.47$ for impact, compared with maximum
model values of $0.52$ and $0.39$, respectively. These averages hide important
differences across stimuli: in one red-cube impact condition, the recorded
\textsc{no-motion} continuation receives human support $0.27$, whereas
Cosmos-3 and Runway each receive $0.40$, showing that a model can disagree with
the recorded future while still matching human expectations. The most
human-supported model also differs by anchor: Veo and Runway receive the
highest support after release, whereas Cosmos-3 does so after impact.
Cosmos-Predict-2.5 is an extreme release case ($0.01$), as its frequent
\textsc{no-motion} continuations were rarely anticipated. Because responses are
open-ended, these scores reflect agreement in the first consequence explicitly
selected by each participant; responses such as ``fall'', ``contact'', and
``rebound'' may refer to different points of a largely shared physical sequence
rather than entirely different predicted futures.

\begin{table}[t]
\centering
\small
\begin{tabular}{lcc}
\toprule
Source & Release & Impact \\
\midrule
Recorded GT         & \textbf{0.61} & \textbf{0.47} \\
Veo                 & 0.52 & 0.30 \\
Runway              & 0.47 & 0.24 \\
Cosmos-3            & 0.34 & 0.39 \\
MAGI-1              & 0.12 & 0.14 \\
PhyWorld            & 0.12 & 0.11 \\
Cosmos-Predict-2.5  & 0.01 & 0.17 \\
\bottomrule
\end{tabular}
\caption{Mean human support for the first physical consequence produced by each
model and by the recorded future, $\mathbb{E}_s[p_H(c\mid s)]$, over $11$
release and $9$ impact stimuli with $15$ human predictions per stimulus.}
\label{tab:human-support}
\end{table}

\paragraph{DRAW: Does the model produce a trajectory?}
Table~\ref{tab:trajectory-production} reports whether each generated
continuation contains a non-degenerate trajectory in the event-defined
\emph{DRAW} window, with static outputs retained as failures rather than
discarded before geometric comparison. Cosmos-Predict-2.5 produces no measurable
trajectory in any of its $19$ evaluable windows, while MAGI-1 does so in only
$2$ of $20$. Discarding these cases would therefore introduce severe
survivorship bias; we treat trajectory production as the primary \emph{DRAW}
result and evaluate geometry only conditional on measurable motion.

\begin{table}[t]
\centering
\small
\begin{tabular}{lcc}
\toprule
Model & Release & Impact \\
\midrule
Cosmos-3           & 6/11 & 7/9 \\
Veo                & 7/11 & 3/9 \\
Runway             & 4/11 & 4/9 \\
PhyWorld           & 1/11 & 1/8 \\
MAGI-1             & 2/11 & 0/9 \\
Cosmos-Predict-2.5 & 0/10 & 0/9 \\
\bottomrule
\end{tabular}
\caption{Production of a measurable trajectory within the event-defined
\emph{DRAW} window. Smaller denominators indicate continuations that were not
evaluable for the corresponding model.}
\label{tab:trajectory-production}
\end{table}

\paragraph{DRAW: How does the trajectory compare?}
For measurable trajectories, $H$--$H$ measures variability among human
drawings, while $H$--$GT$ measures human agreement with the recorded future;
these provide references for the model--human ($M$--$H$) and
model--ground-truth ($M$--$GT$) comparisons. Median Fr\'echet distances are
$D(H,H)=0.86d_0$ for release and $1.79d_0$ for impact, while
$D(H,GT)=0.75d_0$ for release and $1.53d_0$ for impact, showing that the
recorded trajectory is at least as close to humans as human drawings are to one
another. We report model distances only when at least five measurable
trajectories are available. For release, Veo yields $D(M,H)=1.19d_0$ and
$D(M,GT)=0.40d_0$ ($n=7$), while Cosmos-3 yields $D(M,H)=2.23d_0$ and
$D(M,GT)=1.61d_0$ ($n=6$). For impact, Cosmos-3 yields
$D(M,H)=2.22d_0$ and $D(M,GT)=1.58d_0$ ($n=5$). Thus, conditional on
producing motion, Veo's release trajectories lie closer to the recorded future
than Cosmos-3's, while Cosmos-3 remains farther from both humans and the
recorded trajectory. These results should be interpreted together with the
trajectory-production rates above.

\section{Conclusion}

We introduced an event-anchored evaluation of physical foresight that separates
whether a model produces the consequence of an observed event from when and how
that consequence is physically realized. Across six models, we observe distinct
failure stages: some frequently fail to initiate the implied consequence, others
produce it only after a substantial temporal delay, and physically plausible
timing does not necessarily imply physically consistent motion. Human predictions
further show that these failures cannot be explained solely by insufficient
evidence: from even a single event-anchored frame, participants often anticipate
the expected consequence while also revealing genuine ambiguity among plausible
futures. Together, these findings suggest that physical prediction should be
evaluated as a sequence of separable capabilities---consequence production,
temporal anchoring, and physical realization---rather than reduced to a single
similarity or physics score.
{
    \small
    \bibliographystyle{ieeenat_fullname}
    \bibliography{main}
}

\end{document}